# Tiny Graph Neural Networks for Radio Resource Management

Ahmad Ghasemi
University of Massachusetts Amherst
Amherst, MA, USA
aghasemi@umass.edu

Hossein Pishro-Nik
University of Massachusetts Amherst
Amherst, MA, USA
pishro@umass.edu

## ABSTRACT

The surge in demand for efficient radio resource management has necessitated the development of sophisticated yet compact neural network architectures. In this paper, we introduce a novel approach to Graph Neural Networks (GNNs) tailored for radio resource management by presenting a new architecture: the Low Rank Message Passing Graph Neural Network (LR-MPGNN). The cornerstone of LR-MPGNN is the implementation of a low-rank approximation technique that substitutes the conventional linear layers with their low-rank counterparts. This innovative design significantly reduces the model size and the number of parameters. We evaluate the performance of the proposed LR-MPGNN model based on several key metrics: model size, number of parameters, weighted sum rate of the communication system, and the distribution of eigenvalues of weight matrices. Our extensive evaluations demonstrate that the LR-MPGNN model achieves a sixtyfold decrease in model size, and the number of model parameters can be reduced by up to 98%. Performance-wise, the LR-MPGNN demonstrates robustness with a marginal 2% reduction in the best-case scenario in the normalized weighted sum rate compared to the original MPGNN model. Additionally, the distribution of eigenvalues of the weight matrices in the LR-MPGNN model is more uniform and spans a wider range, suggesting a strategic redistribution of weights.

## 1 INTRODUCTION

Efficient radio resource management is pivotal in the ever-evolving landscape of wireless networks, yet it grapples with the challenges of real-time problem-solving due to inherent non-convexity and computational complexity. Traditional optimization techniques often fall short in addressing the scalability and complexity of large-scale, non-convex problems. Motivated by deep learning's successes, Graph Neural Networks (GNNs) have become a key approach for tackling complex wireless network challenges [7, 8]. Existing deep learning solutions for wireless networks, while successful to a degree, are encumbered by their considerable model size and computational intensity, limiting their practicality in real-time scenarios and environments with constrained computational resources [10]. Moreover, these methods exhibit deficiencies in adaptability to the dynamic and ever-changing scale of wireless networks, necessitating a more versatile and scalable solution. Efforts in Tiny GNNs [6, 9] aim to address this, such as using neighbor distillation strategies for implicit knowledge learning from deeper GNNs [9], which depends heavily on the teacher model's quality. Another approach is the Topologically Consistent Magnitude Pruning [6], which maintains topological consistency in extracted subnetworks but complicates optimization and requires extensive experimentation. Despite these innovations, none have been applied to radio resource management. To address this, we introduce the Low Rank Message Passing Graph Neural Network (LR-MPGNN), a novel adaptation of GNNs for the task of radio resource management in multi-user Multi-Input Single-Output (MISO) wireless networks. The LR-MPGNN utilizes a low-rank approximation (LRA) technique to revolutionize GNNs into a compact and efficient paradigm, making it ideal for environments where computational resources are limited.

Conventional deep learning architectures such as Multi-Layer Perceptrons (MLPs) and Convolutional Neural Networks (CNNs) are hampered by scalability and generalization constraints, particularly in expansive wireless network settings. To transcend these limitations, the application of Tiny Machine Learning (Tiny ML) principles, particularly via LRA, revolutionizes GNNs into a paradigm that is both compact and efficient, making it well-suited for deployment in environments where resources are constrained. This transformation results in a Tiny GNN architecture that is significantly more manageable in terms of computational resources. This addresses the scalability and generalization challenges that previous models have encountered.

### 1.1 Contributions

The main contributions of this paper are as follows:

(1) We present LR-MPGNN, an innovative adaptation of GNNs for radio resource management. By integrating LRA technique, we significantly reduce the computational complexity and model size, making our approach ideal for deployment in environments with limited computational resources.
(2) The LR-MPGNN model demonstrates a drastic reduction in model size without significantly compromising performance. Specifically, we achieve a sixtyfold decrease in model size and a reduction of up to 98% in the number of model parameters, facilitating deployment in resource-constrained settings.

(3) By employing TinyML principles and LRA within the GNN framework, our work addresses significant challenges in radio resource management, including computational complexity in real-time problem-solving. Our approach provides a scalable and efficient solution for managing radio resources in dense and dynamic wireless networks.

Through these contributions, we aim to bridge the gap between theoretical machine learning models and their practical application in the complex domain of wireless communications, particularly in leveraging the emerging capabilities of TinyML for efficient and scalable radio resource management.

## 1.2 Organization and Notation

The paper is organized as follows: Section 2 presents the system model and problem definition. Section 3 introduces the proposed low-rank approximated GNN. The evaluation of the proposed approaches is in Section 4. Finally, Section 5 concludes the paper.

**Notation:** In this paper, vectors are shown by small bold-italic face letters $\mathbf{a}$ and capital bold-italic face letters $\mathbf{A}$ show matrices. The rank of matrix $\mathbf{A}$ is represented by rank($\mathbf{A}$). $\mathcal{A}$ is a set and $a$ is a scalar. The $i^{\text{th}}$ element and the number of elements of set $\mathcal{A}$, are shown via $\mathcal{A}[i]$ and $|\mathcal{A}|$, respectively. $|a|$ is the magnitude of the complex number $a$. The transpose and Hermitian of a matrix/vector are shown by $(.)^{\text{T}}$, $(.)^{\dagger}$, respectively. $\|.\|_2$ denotes $l_2$-norm of a vector. $\mathbb{D}^{l \times l}$ and $\mathbb{C}^{m \times n}$ represent a diagonal matrix of dimension $l \times l$ and a complex matrix of dimension $m \times n$. The $n^{\text{th}}$ diagonal element of a diagonal matrix $\mathbb{D}$ is denoted by $\mathbb{D}_n$. $\mathbb{R}$ denotes the set of all real numbers. $\mathbf{I}_N$ denotes the $N \times N$ identity matrix. $\mathbf{0}_N$ and $\mathbf{1}_N$ are the $N$-dimensional all-zeros and all-ones vectors, respectively. We use $\mathcal{CN}(\mu, \sigma^2)$ to denote a circularly symmetric complex Gaussian random vector with mean $\mu$ and variance $\sigma^2$. Finally, $P(.)$, $(.)^*$ and $\mathbb{E}(.)$ denote the probability, the optimum value and the expectation, respectively.

## 2 SYSTEM MODEL AND PROBLEM DEFINITION

This work explores a multi-user Multi-Input Single-Output (MISO) wireless network consisting of $N$ active transceiver pairs, denoted by the set $\mathcal{N} = \{1, 2, \ldots, N\}$. Each transmitter (TX) is equipped with $N_t$ antennas, while receivers (RX) are single-antenna systems, as depicted in Fig. 1.

Considering $\{s_n\}_{n=1}^N$ as the unit-norm signals transmitted from the $n^{\text{th}}$ TX to its corresponding RX, and defining the beamforming/precoding matrix $\mathbf{Q} = [\mathbf{q}_1, \mathbf{q}_2, \ldots, \mathbf{q}_N]^{\text{T}} \in \mathbb{C}^{N \times N_t}$ where $\mathbf{q}_n$ represents the precoder at the $n^{\text{th}}$ transmitter, the received signal at the $n^{\text{th}}$ RX can be modeled as $y_n = \mathbf{h}_{n,n}^{\dagger} \mathbf{q}_n s_n + \sum_{i=1, i \neq n}^{N} \mathbf{h}_{i,n}^{\dagger} \mathbf{q}_i s_i + n_n$, where $\mathbf{h}_{i,n} \in \mathbb{C}^{N_t}$ is the channel vector from the $i^{\text{th}}$ TX to the $n^{\text{th}}$ RX, and $n_n \sim \mathcal{CN}(0, \sigma_n^2)$ denotes the Additive White Gaussian Noise (AWGN) at the $n^{\text{th}}$ RX.

Furthermore, the entire network's channel characteristics are encapsulated in a tensor $\mathbf{H} \in \mathbb{C}^{|\mathcal{V}| \times |\mathcal{V}| \times N_t}$. The elements of this tensor, $\mathbf{H}_{i,n,:} = \mathbf{h}_{i,n} \in \mathbb{C}^{N_t}$ for $\{i, n\} \in \mathcal{N}$, include both diagonal elements (desired channels) and non-diagonal elements (interference channels) for each transceiver pair. This channel tensor is accessible to the central processing unit (CPU). The CPU's responsibility includes the construction and periodic updating of the Deep Learning (DL) model.

## 2.1 Graph Modeling of P2P Wireless Communications

The P2P wireless network under consideration is modeled as a *directed graph*, depicted in Fig. 1. In this graph, each transceiver pair is represented as a vertex, specifically the $n^{\text{th}}$ transceiver pair corresponds to the $n^{\text{th}}$ vertex. Vertex features encapsulate transceiver properties, while a directed edge from vertex $i$ to vertex $j$ indicates interference from TX $i$ to RX $j$, with edge features describing the interference channel properties. Interference is considered only when the TX-RX distance is less than a threshold $T_d$.

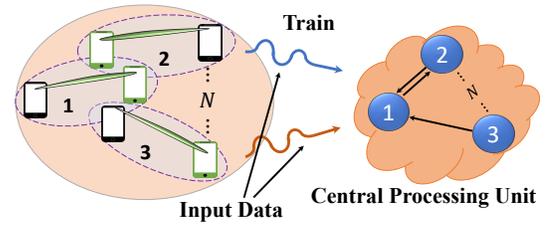

**Figure 1: System Model**

Formally, the graph is denoted as $\mathcal{G}(\mathcal{V}, \mathcal{E})$, with $\mathcal{V}$ and $\mathcal{E}$ representing vertices and edges, respectively. The vertex feature matrix is $\mathbf{Z} \in \mathbb{C}^{|\mathcal{V}| \times (N_t + 2)}$, where $\mathbf{Z}_{n,:} = [\mathbf{h}_{n,n}, w_n, \sigma_n^2]^{\text{T}}$ includes channel vector $\mathbf{h}_{n,n}$, weight $w_n$, and noise power $\sigma_n^2$ for each vertex $n$. The adjacency feature tensor $\mathbf{A} \in \mathbb{C}^{|\mathcal{V}| \times |\mathcal{V}| \times N_t}$ is defined below, where $\mathbf{h}_{i,n}$ representing the channel vector from the $i^{\text{th}}$ TX to the $n^{\text{th}}$ RX.

$$\mathbf{A}_{i,n,:} = \begin{cases} \mathbf{0}_{N_t}, & \text{if } \{i, n\} \notin \mathcal{E}, \\ \mathbf{h}_{i,n}, & \text{otherwise.} \end{cases} \quad (1)$$

Using these definitions, the received signal at the $n^{\text{th}}$ RX is reformulated as $y_n = \mathbf{Z}_{n,1:N_t}^{\dagger} \mathbf{q}_n s_n + \sum_{i=0, i \neq n}^{N} \mathbf{A}_{i,n,:}^{\dagger} \mathbf{q}_i s_i + n_n$, leading to the Signal-to-Interference-plus-Noise Ratio (SINR) at the $n^{\text{th}}$ RX as:

$$\text{SINR}_n = \frac{|\mathbf{Z}_{n,1:N_t}^{\dagger} \mathbf{q}_n|^2}{\sum_{i=1, i \neq n}^{N} |\mathbf{A}_{i,n,:}^{\dagger} \mathbf{q}_i|^2 + \mathbf{Z}_{n,N_t+2}}. \quad (2)$$

Given this SINR, the objective of the system is to find the optimal beamformer that maximizes the weighted sum rate. The problem is formulated as:

$$\max_{\mathbf{Q}} \quad \sum_{n \in \mathcal{N}} \mathbf{Z}_{n,N_t+1} \log_2(1 + \text{SINR}_n), \quad (3a)$$

$$\text{s.t.} \quad \|\mathbf{q}_n\|_2^2 \leq P_{\max}, \ \forall \ n \in \mathcal{N} \quad (3b)$$

where $\mathbf{Z}_{n,N_t+1}$ represents the weight for the $n^{\text{th}}$ pair, based on the definition of the vertex feature matrix $\mathbf{Z}$.

To optimize this system, we utilize a three-layer message passing graph neural networks (MPGNN) described in [3, 4, 7], in which in each layer, each vertex updates its representation by aggregating features from its neighbor vertices. Specifically, the $n^{\text{th}}, n \in \mathcal{N}$, vertex in the $k^{\text{th}}, k \in \{1, 2, 3\}$, layer of MPGNN updates by:

$$y_n^{(k)} = \text{MLP2}\left(x_n^{(k-1)}, \max_{j \in B(n)} \left\{\text{MLP1}\left(x_j^{(k-1)}, \mathbf{A}_{j,n,:}\right)\right\}\right),$$
$$x_n^{(k)} = \beta\left(y_n^{(k)}\right), \quad (4)$$

where MLP1 and MLP2 are two different multi-layer perceptrons (MLPs). In addition, $x_n^{(0)} = \mathbf{Z}_{n,:}$ is the input feature of node $n$, $B(n)$ denotes the set of the neighbors of $n$, $\mathbf{A}_{j,n,:}$ represents the edge feature of the edge $(j, n)$, and $\beta$ is the sigmoid function. The output of this optimization problem is the beamforming vector $\mathbf{Q}$ that it is optimized by minimizing the loss function $l_\Theta$ in the final layer:

$$l_\Theta = -\mathbb{E}(\sum_{n=1}^{N} \mathbf{Z}_{n,N_t+1} \log_2(1 + \text{SINR}_n(\Theta))). \quad (5)$$

Our analysis in Section 4 demonstrates that the size of the trained MPGNN model scales with the number of transmit (TX) antenna elements, denoted by $N_t$. The model input $\mathbf{Z}_{n,:}$ is dependent on this parameter. From a wireless communication engineering perspective, a larger number of antenna elements is preferred as it enhances the communication's system performance [1, 2]. However, when considering deployment for edge or on-device implementation, a smaller model size is advantageous. To reconcile these opposing requirements and reduce the model size, we employ low-rank factorization discussed in the subsequent sections. Henceforth, we refer to the unaltered model as the *original model*.

## 3 MODEL SIZE REDUCTION VIA LOW-RANK LINEAR LAYERS

In the proposed MPGNN model in this section, we aim to reduce the model size without significantly compromising the learning capability and the performance of the communication system. This is achieved by substituting standard linear layers of MPGNN with custom low-rank linear layers. These layers offer a parameter-efficient alternative to traditional dense layers, particularly beneficial in large-scale models.

### 3.1 Low-Rank Linear Layer

The core idea behind the low-rank linear layer is to decompose a typical linear operation into two sequential linear transformations involving lower-rank matrices. Given an input feature vector $\mathbf{x} \in \mathbb{R}^{d_{\text{in}}}$, where $d_{\text{in}}$ represents the input dimension, the output $\mathbf{y} \in \mathbb{R}^{d_{\text{out}}}$ of a standard linear layer (with an output dimension $d_{\text{out}}$) is computed as:

$$\mathbf{y} = \mathbf{Wx} + \mathbf{b}, \quad (6)$$

where $\mathbf{W} \in \mathbb{R}^{d_{\text{out}} \times d_{\text{in}}}$ is the weight matrix and $\mathbf{b} \in \mathbb{R}^{d_{\text{out}}}$ is the bias vector. In contrast, the low-rank linear layer decomposes $\mathbf{W}$ into two matrices $\mathbf{U} \in \mathbb{R}^{d_{\text{in}} \times r}$ and $\mathbf{V} \in \mathbb{R}^{r \times d_{\text{out}}}$, where $r$ is the rank of the approximation ($r \ll \min(d_{\text{in}}, d_{\text{out}})$). The operation thus becomes:

$$\mathbf{y} = (\mathbf{VU})\mathbf{x} + \mathbf{b} \quad (7)$$

This decomposition cuts the number of parameters from $d_{\text{in}} \times d_{\text{out}}$ to $r \times (d_{\text{in}} + d_{\text{out}})$, leading to a more compact model.

We can reduce the number of parameters of the system so long as the number of parameters $r \times (d_{\text{in}} + d_{\text{out}})$ is less than A (i.e., mn). If we would like to reduce the number of parameters in A by a fraction p, we require the following to hold.

### 3.2 Implementation in LR-MPGNN

In the context of our MPGNN model, the standard linear layers are replaced with the proposed low-rank linear layers. Algorithm 1 outlines this process:

---
**Algorithm 1** Low-Rank Linear Layer in MPGNN
---
**Require:** Input features $\mathbf{X} \in \mathbb{R}^{N \times d_{\text{in}}}$, rank $r$
**Ensure:** Output features $\mathbf{Y} \in \mathbb{R}^{N \times d_{\text{out}}}$
1: Initialize $\mathbf{U} \in \mathbb{R}^{d_{\text{in}} \times r}, \mathbf{V} \in \mathbb{R}^{r \times d_{\text{out}}}$
2: **for** each layer in the MPGNN **do**
3:     Compute $\mathbf{Y} \leftarrow (\mathbf{VU})\mathbf{X}$
4:     Apply activation function (e.g., ReLU) to $\mathbf{Y}$
5:     $\mathbf{X} \leftarrow \mathbf{Y}$      ▷ Feed to next layer
6: **end for**
7: **return** $\mathbf{Y}$
---

### 3.3 Impact of Rank $r$ on Model Size and System Performance

The choice of rank $r$ in the low-rank linear layer is pivotal, as it directly influences the balance between model complexity and the communication system performance. The minimum value of $r$, typically set to 1, offers the most significant reduction in model size. This extreme compression, however, may lead to substantial information loss, adversely affecting the model's representational capacity and the system's performance. Conversely, the maximum value of $r$, equal to the minimum of the input and output dimensions ($\min(d_{\text{in}}, d_{\text{out}})$), represents no reduction in rank and hence no compression. This setting retains the full capacity of the original linear layer but offers no advantages in terms of model size reduction.

In practice, the optimal value of $r$ is found between these two extremes. A smaller $r$ results in a more compact model, beneficial for deployment in resource-constrained environments, but it may compromise the model's ability to capture complex patterns in data. On the other hand, a larger $r$ preserves more information and may yield better system's performance, but with diminishing returns in terms of model size reduction and computational efficiency. Therefore, selecting an appropriate $r$ involves balancing the trade-off between the model size and the communication system performance, often requiring empirical experimentation and validation on specific tasks and datasets.

### 3.4 Parameter Reduction in Graph Neural Networks

GNNs are powerful tools for learning on graph-structured data. However, the complexity of these models often leads to a large number of parameters, which can be a hindrance for deployment on resource-constrained devices. Low-rank matrix factorization is a technique employed to reduce the number of parameters in neural networks, thereby decreasing the computational cost and memory requirements. In this section, we derive a general formula for the parameter reduction fraction $p$ after applying low-rank approximations to the weight matrices in a GNN.

*3.4.1 General Formula for Parameter Reduction.* Consider a GNN with two fully connected layers denoted as MLP1 and MLP2. The first layer, MLP1, has dimensions $[l_{11} \cdot N_t, l_{12}, l_{13}]$, and the second layer, MLP2, follows with dimensions $[l_{13} + l_{21} \cdot N_t, l_{22}, l_{23} \cdot N_t]$. The original number of parameters for each layer is given by the product of its dimensions. After applying low-rank approximations, MLP1 and MLP2 are factorized into pairs of matrices with ranks $a_1$ and $a_2$ respectively. The new number of parameters for each layer is thus the sum of the parameters of these factorized matrices.

The parameter reduction fraction $p$ is then defined as:

$$p = 1 - \frac{\text{Low-Rank Parameters}}{\text{Original Parameters}} \quad (8)$$

where Low-Rank Parameters is the sum of the parameters after low-rank approximation for both layers, and Original Parameters is the sum of the parameters before approximation.

*3.4.2 Our Case.* Applying the general formula to our case where MLP1 has dimensions $[6 \cdot N_t, 64, 64]$ and MLP2 has dimensions $[64 + 4 \cdot N_t, 512, 2 \cdot N_t]$, we map the layer dimensions to our general variables $l_{11}, l_{12}, l_{13}, l_{21}, l_{22}, l_{23}$ as follows: $l_{11} = 6, l_{12} = l_{13} = 64, l_{21} = 4, l_{22} = 512, l_{23} = 2$.

Substituting these values into the general formula for $p$, we obtain the specific expression:

$$p = \frac{-3N_t a_1 - 3N_t a_2 + 1728 N_t - 96 a_1 - 544 a_2 + 18432}{576(3N_t + 32)} \quad (9)$$

This expression allows us to calculate the parameter reduction for any given rank approximations $a_1$ and $a_2$, and the number of antennas $N_t$ in the model. In Subsection 4.3, we will demonstrate how the $p$ value varies with different values of $N_t$, $a_1$, and $a_2$.

## 3.5 Adaptability and Hardware Efficiency

The design and implementation of LR-MPGNN consider not only computational efficiency and scalability but also adaptability to dynamic network environments and efficient hardware resource utilization.

*3.5.1 Hardware Characteristics and Resource Utilization.* The LR-MPGNN model is specifically designed to operate within the constraints of hardware commonly used in wireless network systems. The LRA technique significantly reduces the computational complexity and memory requirements, making the LR-MPGNN model suitable for deployment on devices with limited computational power and memory, such as IoT devices and edge computing nodes.

*3.5.2 Operational Dynamics and Adaptability.* The LR-MPGNN model is able to adapt to changes within the network environment effectively. This adaptability is crucial for managing radio resources in dynamic and dense wireless networks where network conditions can fluctuate rapidly. While the primary training of the LR-MPGNN model occurs offline, leveraging historical data and simulations to capture a wide range of network scenarios, the model is also equipped with mechanisms for incremental learning. This enables the LR-MPGNN to update its parameters in response to new environmental conditions or network configurations without requiring a complete retraining process.

The decision to retrain the model depends on the extent of environmental or network changes. For significant shifts in network topology or usage patterns, a more comprehensive retraining may be warranted. However, for minor changes, the model can adjust through lighter updates, ensuring continuous optimization of radio resources without substantial computational overhead.

## 4 PERFORMANCE ANALYSIS

This section assesses the proposed tiny MPGNN models, namely LR-MPGNN. It focuses on assessing the impact of low-rank approximation on several key aspects: the size of the model, the performance of the communication system, and the distribution of the model's weights.

For our dataset, we simulated transceiver pairs within a specified rectangular area, randomly placing transmitters and distributing their corresponding receivers uniformly within a distance range of $[d_{\min}, d_{\max}]$. The channel models, based on the approach in [3], define the TX-RX channel as $\mathbf{h}_{j,i} = 10^{-L(d_{ji})/20} \sqrt{\psi_{ji} \rho_{ji}} \mathbf{g}_{ji}$ for all pairs in $\mathcal{N}$. Here, $L(d_{ji}) = 148.1 + 37.6 \log_2(d_{ji})$ represents path-loss at distance $d_{ji}$ (in kilometers), $\psi_{ji}$ is the antenna gain (9 dBi), $\rho_{ji}$ is the shadowing coefficient, and $\mathbf{g}_{ji}$ follows a $\mathcal{CN}(\mathbf{0}_{N_t}, \mathbf{I}_{N_t})$ small-scale fading distribution.

To reduce Channel State Information (CSI) training overhead, we assumed channels exist only for transceiver pairs separated by less than 500 meters. Our dataset split comprises 2000 training samples and 500 testing samples, each including $N$ transceiver pairs.

The employed GNN architecture, identical to those in [2, 7], is a 3-layer graph neural network, as detailed in Section 2.1. It inputs channel states $\{\mathbf{Z}^\dagger_{n,1:N_t}\}_{n=1}^N$ and users' weights $\{w_n\}_{n=1}^N$, producing user beamforming vectors. The loss function is as defined in (5). For optimization, we utilized the Adam algorithm [5] with a learning rate of 0.001. The number of transceiver pairs $N$ and Signal-to-Noise Ratio (SNR) settings were consistent across both training and testing phases.

### 4.1 Model Size

This subsection presents a comparison of the relative sizes of low-rank approximated models (LR-MPGNN) to the original model (MPGNN), using variable ranks $a_1$ and $a_2$. This comparison is key to understanding the efficacy of low-rank approximation techniques in reducing model size. Here, the number of antenna elements at the transmitter, denoted as $N_t$, is 512, and the number of transceiver pairs, $N$, is 3. Additionally, the maximum values for $a_1$ and $a_2$ are set to 64 and 512, respectively, because the maximum possible rank for MLP1 is 64, and for MLP2, it is 512.

First, we assess the impact of adjusting the rank parameters $a_1$ and $a_2$ on the model size, as detailed in Table 1. The values presented in this table are calculated by dividing the size of the low-rank approximated models by the size of the original model before the application of low-rank approximation. This table shows that as the rank parameters $a_1$ and $a_2$ increase, the size of the approximated model also grows but remains smaller than the full-sized original model. Particularly noteworthy is the substantial size reduction when both rank parameters are at their minimum (both $a_1$ and $a_2$ set to 4), where the model becomes 59 times smaller than the original, illustrating its potential in scenarios requiring lower storage or computational resources.

**Table 1: Relative Size of Low-Rank Approximated Models Compared to Original Model**

| $a_1$ \ $a_2$ | 4 | 16 | 32 | 64 | 128 | 256 | 512 |
|---|---|---|---|---|---|---|---|
| 4 | 58.82 | 22.42 | 12.29 | 6.45 | 3.31 | 1.68 | 0.84 |
| 16 | 25.87 | 15.10 | 9.71 | 5.66 | 3.09 | 1.62 | 0.83 |
| 32 | 14.81 | 10.51 | 7.58 | 4.87 | 2.84 | 1.55 | 0.81 |
| 64 | 7.98 | 6.54 | 5.27 | 3.80 | 2.44 | 1.42 | 0.77 |

Furthermore, a general trend of increasing relative size of the approximated model is observed with an increase in either $a_1$ for a fixed $a_2$ or vice versa. This indicates that both rank parameters significantly impact the resulting model size. Increasing both $a_1$ and $a_2$ simultaneously results in a closer approximation to the original model size, demonstrating a critical balance between model complexity and approximation fidelity. Additionally, the non-linear nature of size reduction, with unequal contributions from different dimensions, suggests a limit to the achievable compression without significant information loss.

In addition, the values less than one in the last column demonstrate that the size of LR-MPGNN increases when the selected rank for MLP2 is 512. This could be attributed to the fact that the actual rank of MLP2 is less than 512.

### 4.2 Communication System Performance

In this subsection, we evaluate the system performance of the LR-MPGNN model based on the weighted sum rate defined in (3). The results represent the weighted sum rate achieved by LR-MPGNN, normalized by the weighted sum rate of the original MPGNN model.

**Table 2: Normalized weighted sum rate**

| $a_1$ \ $a_2$ | 4 | 16 | 32 | 64 | 128 | 256 | 512 |
|---|---|---|---|---|---|---|---|
| 4 | 0.762 | 0.819 | 0.596 | 0.739 | 0.723 | 0.57 | 0.765 |
| 16 | 0.971 | 0.603 | 0.896 | 0.68 | 0.68 | 0.487 | 0.819 |
| 32 | 0.743 | 0.603 | 0.652 | 0.714 | 0.67 | 0.64 | 0.99 |
| 64 | 0.80 | 0.71 | 0.612 | 0.552 | 0.844 | 0.884 | 0.853 |

Table 2 indicates a varied performance across different parameter settings for $a_1$ and $a_2$. Notably, the model achieves peak performance at $a_1 = 16$ and $a_2 = 4$ with a normalized weighted sum rate of 0.971. This suggests that the LR-MPGNN model is highly effective under these specific settings, aceiving a very close performance to that of the original MPGNN model.

Conversely, the configuration at $a_1 = 16$ and $a_2 = 256$ yields the lowest normalized rate of 0.487, which could be indicative of suboptimal parameter selection for these conditions. The presence of this outlier prompts further investigation into the underlying causes of such a performance dip. We should note that to ensure a fair comparison, we kept all parameters constant, including the optimizer, learning rate, batch size, among others, during the training of both the MPGNN and LR-MPGNN models. Therefore, while a 50% decrease in the performance might seem substantial, it is important to recognize that this could potentially be improved by adjusting the aforementioned parameters.

Interestingly, the model performance does not exhibit a monotonically increasing or decreasing trend with respect to the parameters $a_1$ and $a_2$, suggesting a complex relationship between these parameters and the resulting efficiency. For example, a notably high value of 0.99 is observed at $a_1 = 32$ and $a_2 = 512$, which contrasts with the adjacent values and indicates an area of potential optimal parameter space.

These findings underscore the importance of parameter tuning in the application of the LR-MPGNN model and suggest that further studies should be conducted to understand the dynamics of the model's performance over its parameter space fully.

### 4.3 Analysis of Parameter Reduction

To better understand the impact of low-rank approximations on parameter reduction in MPGNN, we visualize the parameter reduction fraction $p$ in (9) as a function of the ranks $a_1$ and $a_2$ used in the approximations. Here, $N_t = 512$ and $N = 3$.

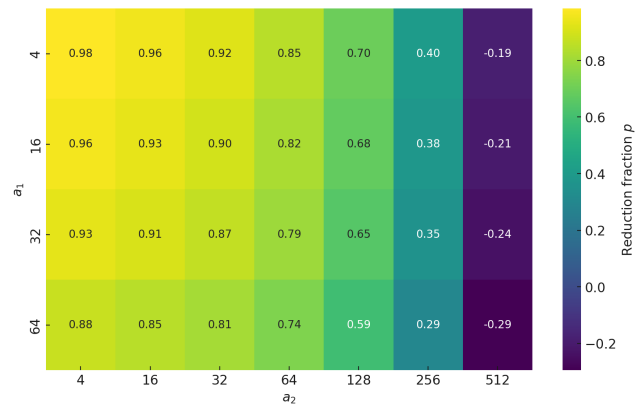

**Figure 2: Heatmap visualization of $p$**

The heatmap in Figure 2 provides a color-coded representation of the reduction fraction $p$, with each cell corresponding to a specific combination of $a_1$ and $a_2$. The y-axis represents the rank $a_1$, and the x-axis represents the rank $a_2$. The color in each cell indicates the value of $p$, following a scale where blue signifies lower values of $p$ (indicating less parameter reduction or an increase in parameters), and yellow represents higher values (indicating more substantial parameter reduction). The annotated values within each cell provide the precise reduction fraction $p$ for each rank combination. This figure provides insightful observations: 1. *Effectiveness of Low-Rank Approximations:* For certain combinations of $a_1$ and $a_2$, the reduction fraction $p$ approaches 1, indicating a significant reduction in parameters, which is desirable for model efficiency, 2. *Impact of Higher Ranks:* As the ranks $a_1$ and $a_2$ increase, the reduction in parameters diminishes, and for some higher values, $p$ even becomes negative. This indicates an increase in the number of parameters, suggesting that high ranks may counteract the benefits of low-rank approximations, and, 3. *Optimal Rank Selection:* The optimal choice of ranks for low-rank approximations depends on the desired balance between model complexity and parameter reduction. Lower ranks generally lead to higher parameter reduction but may also impact the model's ability to learn complex representations.

## 4.4 Comparison of Weight Distributions

This section discusses weight distribution changes in our MPGNN model before and after applying low-rank approximation.

*4.4.1 Original Model Weight Distribution.* The weight distribution of the original MPGNN model is presented in Figure 3. The distribution exhibits a Gaussian-like behavior centered around zero. This is a typical characteristic of well-initialized neural networks where the weights are often sampled from a distribution with zero mean, which promotes balanced learning dynamics and prevents the early saturation of neurons' activation functions.

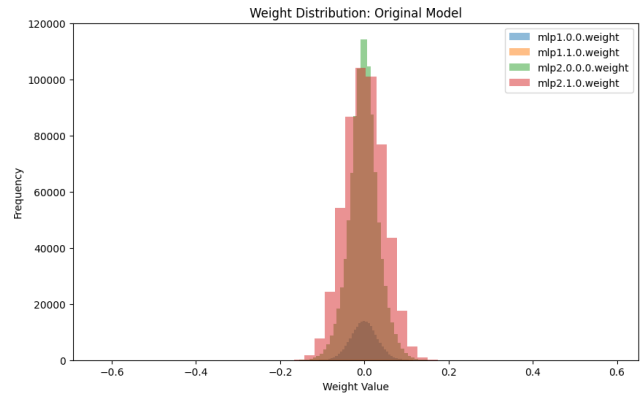

**Figure 3: Weight distribution of the original GNN model before low-rank approximation. The weights are normally distributed, indicating a standard initialization scheme.**

*4.4.2 Low-Rank Approximated Model Weight Distribution.* After applying low-rank approximation, the weight distribution undergoes a substantial transformation, as depicted in Figure 4. The resulting distribution is more uniform and spans a wider range, suggesting a redistribution of weights towards a more diversified set of values. This redistribution magnifies as ranks $a_1$ and $a_2$ decrease. This can be attributed to the factorization process, which decomposes the weight matrices into lower-dimensional spaces, thereby altering their inherent structure and potentially leading to a broader exploration of the solution space during training.

*4.4.3 Discussion.* A comparative analysis of Figures 3 and 4 indicates that low-rank approximation not only reduces the model's complexity by decreasing the number of parameters but also impacts the weights' distribution. One potential advantage of such a change could be the introduction of regularization effects, as the model is compelled to maintain the communication system's performance with a constrained set of parameters, possibly leading to better generalization. However, this alteration also raises concerns regarding the model's capacity to represent complex functions, as the expressiveness of a neural network is partially determined by its weight diversity. Thus, it is crucial to carefully choose the rank of approximation to strike a balance between model efficiency and representational power.

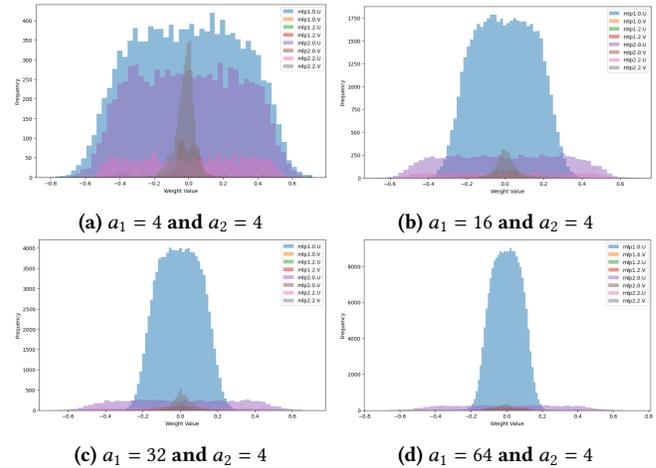

(a) $a_1 = 4$ and $a_2 = 4$    (b) $a_1 = 16$ and $a_2 = 4$

(c) $a_1 = 32$ and $a_2 = 4$    (d) $a_1 = 64$ and $a_2 = 4$

**Figure 4: Weight distribution of LR-MPGNN.**

## 5 CONCLUSION

In conclusion, this work has successfully demonstrated the viability of leveraging low-rank approximation within the architecture of Graph Neural Networks for radio resource management. The proposed Tiny Message Passing Graph Neural Network (TMP-GNN) stands as a testament to the efficiency of model compactness without significant performance compromise. Our results are twofold: they reveal a substantial reduction in model size by a factor of 60 and a decrease in the number of model parameters by 98%, which is a remarkable feat in neural network optimization.

Despite the considerable reduction in model complexity, the performance metrics of the LR-MPGNN model offer compelling evidence of its efficacy. In the best-case scenario, the performance degradation compared to the original MPGNN model is a negligible 2%, while the worst-case scenario shows a 50% decrease. This delineates the conditions under which the TMP-GNN model maintains high efficiency, providing valuable insights for its deployment in various scenarios.

Moreover, the analysis of the eigenvalue distribution for the weight matrices in the LR-MPGNN model indicates a uniform spread across a wider range. This suggests an advantageous redistribution of weights that underpins a more diversified representation capability, potentially enhancing the model's ability to generalize and thus further justifying the low-rank approach.

Overall, our research underlines the potential of low-rank approximations in reducing the computational overhead of GNNs, while retaining a robust performance profile. These findings not only pave the way for more efficient neural network designs in the field of radio resource management but also open avenues for future research in model optimization strategies across various domains.

## ACKNOWLEDGEMENTS
This work was partially supported by NSF under Grant CNS-2150832.